\documentclass[runningheads]{llncs}
\usepackage[T1]{fontenc}
\usepackage{tabularray}
\usepackage{booktabs}
\usepackage[export]{adjustbox}
\usepackage{capt-of}
\usepackage{xparse}
\usepackage{comment}
\usepackage{graphicx}
\usepackage{amsmath}
\usepackage{mathtools}
\usepackage{amssymb}
\usepackage{multirow}
\usepackage[dvipsnames]{xcolor}
\usepackage{pifont}% http://ctan.org/pkg/pifont
\usepackage{subcaption}
\usepackage{hyperref}       % hyperlinks
\usepackage{url}            % simple URL typesetting

\usepackage{algorithm} %added for pseudocode
\usepackage{algpseudocode} %added for pseudocode

\newcommand{\xmark}{\ding{55}}%
\newcommand{\greencheck}{\ifmmode\text{\textcolor{BurntOrange}{\ding{51}}}\else\textcolor{BurntOrange}{\ding{51}}\fi}

\newcommand{\yellowcheck}{\ifmmode\text{\textcolor{ForestGreen}{\ding{51}}}\else\textcolor{ForestGreen}{\ding{51}}\fi}

\newcommand{\littletaller}{\mathchoice{\vphantom{\big|}}{}{}{}}
\newcommand{\restr}[2]{{% we make the whole thing an ordinary symbol
  \left.\kern-\nulldelimiterspace% automatically resize the bar with \right
  #1% the function
  \littletaller% pretend it's a little taller at normal size
  \right|_{#2}% this is the delimiter
  }}
\newcommand*\diff{\mathop{}\!\mathrm{d}}
\newcommand{\MAP}{\textsc{map}}

\begin{document}
\title{Laplacian Segmentation Networks Improve Epistemic Uncertainty Quantification}
\titlerunning{Laplacian Segmentation Networks}
\author{Kilian Zepf\inst{1*}  \and
Selma Wanna\inst{2*} \and
Marco Miani\inst{1} \and
Juston Moore\inst{2} \and
Jes Frellsen\inst{1} \and
Søren Hauberg\inst{1} \and
Frederik Warburg\inst{3} \and
Aasa Feragen\inst{1} }

% index{Zepf, Kilian}
% index{Wanna, Selma}
% index{Miani, Marco}
% index{Moore, Juston}
% index{Frellsen, Jes}
% index{Hauberg, Søren}
% index{Warburg, Frederik}
% index{Feragen, Aasa}

\authorrunning{K. Zepf et al.}
\institute{Technical University of Denmark, Kongens Lyngby, Denmark \email{\{kmze,mmia,jefr,sohau,afhar\}@dtu.dk} \and
Los Alamos National Laboratory, Los Alamos, USA \email{\{slwanna,jmoore01\}@lanl.gov} \and
Teton.ai, Copenhagen, Denmark \email{frederik@teton.ai}\\
* denotes equal contribution
}
\maketitle              % typeset the header of the contribution
\begin{abstract}
Image segmentation relies heavily on neural networks which are known to be overconfident, especially when making predictions on out-of-distribution (OOD) images. This is a common scenario in the medical domain due to variations in equipment, acquisition sites, or image corruptions. This work addresses the challenge of OOD detection by proposing Laplacian Segmentation Networks (LSN): methods which jointly model epistemic (model) and aleatoric (data) uncertainty for OOD detection. In doing so, we propose the first Laplace approximation of the weight posterior that scales to large neural networks with skip connections that have high-dimensional outputs. We demonstrate on three datasets that the LSN-modeled parameter distributions, in combination with suitable uncertainty measures, gives superior OOD detection.

\end{abstract}

\section{Introduction}
Segmentation is extensively used to quantify organs or anomalies and highlight important image features to clinicians. Its widespread application necessitates strict requirements for safe and interpretable operation. However, modern approaches rely on neural networks which are infamously overconfident on predictions outside of their training distributions \cite{hendrycks2016baseline}.  As a result, downstream predictions may be confidently incorrect despite their high accuracy on in-distribution (ID) data, rendering every following analysis based on the prediction unreliable. Thus, detecting gradual distribution shifts is crucial for medical imaging tasks.

\textbf{In this work}, we study Laplace approximations (LA) for epistemic uncertainty quantification in binary image segmentation models. Current Laplace approximations \cite{daxberger2021laplaceredux} scale quadratically with the output dimension of the neural network, which prevent their usage in segmentation. We develop a fast Hessian approximation for deep architectures with skip connections, which are integral components of segmentation networks, e.g., U-net \cite{ronneberger2015}. This enables us to combine the aleatoric logit distribution of Stochastic Segmentation Networks (SSN) \cite{monteiro2020} with Laplace approximations for epistemic uncertainty quantification.

We leverage aspects of the ValUES framework \cite{kahl2024values} to measure the effects of uncertainty estimation on OOD detection. We investigate how useful the inferred uncertainties are for OOD detection and how well LSNs can separate aleatoric and epistemic uncertainty. On three medical binary segmentation tasks, we show that the proposed method provides competitive outlier classification performance and assigns higher uncertainty to OOD datasets. The code is available.\footnote{\href{https://github.com/kilianzepf/laplacian_segmentation}{https://github.com/kilianzepf/laplacian\_segmentation}}

\section{Background}
While several different sources and taxonomies of uncertainties have been proposed \cite{gawlikowski2021,ditlevsen2008}, the Bayesian framework \cite{bishop2006,kendall2017} distinguishes between two of them: aleatoric and epistemic. These can be derived directly from the Bayesian model average (BMA) predictive distribution
\begin{equation}
  \label{pred_dist}
   p(y \vert x, D) = \int \underbrace{p(y \vert x,\theta)}_\text{likelihood} \underbrace{p(\theta \vert D)}_\text{posterior} \diff \theta,
\end{equation}
where $(x, y)$ is an input-output pair, $D$ is the training data and $\theta$ the model parameters. The Shannon entropy $\text{H}(\cdot)$ of the predictive distribution is a common measure of total predictive uncertainty \cite{houlsby2011bayesian}. This \emph{predictive entropy} can be decomposed into the \emph{expected entropy} as a measure of aleatoric uncertainty, and the \emph{mutual information}, $\mathbf{I}$, representing epistemic uncertainty \cite{kendall2017}: 
\begin{equation}
  \label{uncertainty_mi}
   \underbrace{\text{H}(p(y \vert x, D))}_\text{predictive entropy} =  \underbrace{\mathbb{E}_{p(\theta \vert D)}[\text{H}(p(y \vert x,\theta))]}_\text{expected entropy - aleatoric} +  \underbrace{\mathbf{I}[p(y, \theta \vert x,D)]}_\text{mutual information - epistemic} \text{\cite{schweighofer2023}}. 
\end{equation}

Aleatoric uncertainty represents noise or variations that arise from ambiguities in the data, e.g., vague tumor boundaries resulting from gradual tissue infiltration. Epistemic uncertainty, in this framework measured as mutual information $\mathbf{I}$, quantifies the degree to which the model itself should be trusted. However, recent works  
caution against using this decomposition by demonstrating behavioral incoherences of mutual information \cite{schweighofer2023,schweighofer2023quantification,pmlr-v216-wimmer23a}, calling for new formulations to calculate epistemic uncertainty. Table \ref{table_measures} lists the measures included in our study, such as expected pairwise KL-divergence (EPKL) \cite{schweighofer2023}: a recently proposed method which uses pairwise comparisons between the predictive distributions of possible models and weights to calculate epistemic uncertainty. We also consider the pixel-wise variance of mean predictions averaged over samples from the posterior distribution, a measure we call Pixel Variance. Prior work \cite{kahl2024values,mucsanyi2024benchmarking} suggests there is no universal method for uncertainty estimation. 
Thus, the measure of magnitude of a targeted uncertainty type becomes a design choice that needs to be selected for the dataset and task at hand.
\begin{table}
    \begin{center}
    \caption{Overview of uncertainty measures with targeted uncertainty types.}
    \label{table_measures}
    \begin{tabular}{|l|l|l|}
        \hline
        Targeted Type &  Uncertainty Measure & Definition\\
        \hline
        Predictive  &  Predictive Entropy & $H(\mathbb{E}_{q(\theta \vert D)}[p(y \vert x, \theta)])$\\
        Aleatoric  &  Expected Entropy & $\mathbb{E}_{q(\theta \vert D)}[H(p(y \vert x, \theta))]$\\
        Epistemic  & Mutual Information & $I(p(y,\theta \vert x, D))$\\
        Epistemic  & Expected Pairwise KL & $\mathbb{E}_{q(\theta \vert D)}[\mathbb{E}_{q(\Tilde{\theta} \vert D)}[D_{\text{KL}}(p(y \vert x, \theta) \vert \vert p(y \vert x, \Tilde{\theta}) )]]$\\
        Epistemic  & Pixel Variance & $\text{Var}_{q(\theta \vert D)}[\textrm{sigmoid}(\mu_{\theta})]$\\
        \hline
    \end{tabular}
    %\vspace*{-\baselineskip}
    \end{center}
\end{table}

Kendall et al. \cite{kendall2017} recommend jointly modelling aleatoric and epistemic uncertainty for regression and classification tasks in computer vision. Prior methods that model both types of uncertainty typically combine Mean-Variance networks with diagonal covariance matrices and Dropout \cite{kendall2017} or utilize Gaussian-Process based convolutional layers \cite{popescu2021distributional}. Recent works, however, focus on modeling either one or the other component \cite{kohl2018probunet,monteiro2020}. Generally, aleatoric uncertainty techniques rely on mixing deterministic segmentation architectures with generative components \cite{baumgartner2019phiseg,kohl2019hierarchical,selvan2020uncertainty}. Common epistemic modeling techniques incorporate dropout, ensembles and multi-head models \cite{kendall2017,blundell2016ensemble,lee2016stochastic,rupprecht2017learning}.

Fairly evaluating the performance of uncertainty estimation techniques on real-world tasks is challenging due to combinatorial design factors. Kahl et al. \cite{kahl2024values} address this issue by providing a framework which standardizes evaluations for uncertainty estimation methods. Our work coheres to this framework in the sense that we determine how Prediction Models, Uncertainty Measures and Aggregation strategies impact uncertainty estimation on OOD detection tasks for a fixed U-net architecture.
Our results add insight to the recent discussion which calls into question the suitability of common uncertainty measures \cite{schweighofer2023,schweighofer2023quantification,pmlr-v216-wimmer23a}.
\section{The Laplacian Segmentation Networks}
To model the posterior distribution in Eq.~\eqref{pred_dist} we apply Laplace's method which approximates the weight posterior with a Gaussian distribution $q(\theta)$ around a local mode $\theta_{\MAP}$ using the Hessian matrix $\mathbf{H}$ \cite{mackay1992bayesian}
\begin{equation}
   q(\theta)= \mathcal{N}(\theta \vert \theta_{\MAP}, \mathbf{H}^{-1}). 
\end{equation}
Evaluating $\mathbf{H}$ is computationally infeasible because of the quadratic complexity in network parameters and the large output dimensions for segmentation. We improve upon Hessian approximation techniques \cite{daxberger2021laplaceredux,botev2020} by extending recent progress in scaling LA for images \cite{miani_2022_neurips} to segmentation networks with skip connections.
\subsection{Laplace Approximation of the Mean Network}
We can reformulate the integral for the predictive distribution over the binary predictions $y$ in Eq.~\eqref{pred_dist} by integrating over logits $\eta$ to obtain 
\begin{equation}
    \label{pred_dist_with_logits}
 p(y \vert x, D) = \iint p(y \vert \eta) p(\eta \vert x,\theta) p(\theta \vert D) \diff \eta \diff \theta.
\end{equation}
Following \cite{kendall2017} and \cite{monteiro2020}, we model the conditional distribution over logits $p(\eta \vert x,\theta)$ as a normal distribution parametrized by neural networks $\mu$ and $\Sigma$ :
\begin{equation}
    \label{logit_dist}
    \eta \vert x \sim \mathcal{N}(\mu(x, \theta_1),\Sigma(x,\theta_2)),
\end{equation}
and assume pixel-wise independence for the predicted labels given the logits. Thus, we can model $p(y \vert \eta)$ for each pixel $s$ as a Bernoulli distribution parametrized by the sigmoid of the respective logit. 
Since the size of the covariance matrix $\Sigma$ scales quadratically with the number of pixels $S$ in the image, we use the low-rank parameterisation of \cite{monteiro2020}:
\begin{equation}
    \Sigma(x) = D(x) + P(x)^T P(x),
\end{equation}
i.e.~the variance network $\Sigma(x)$ is implemented with two networks $D(x)$ and $P(x)$. 

\begin{figure*}[t]
\begin{center}
\includegraphics[width=0.8\linewidth]{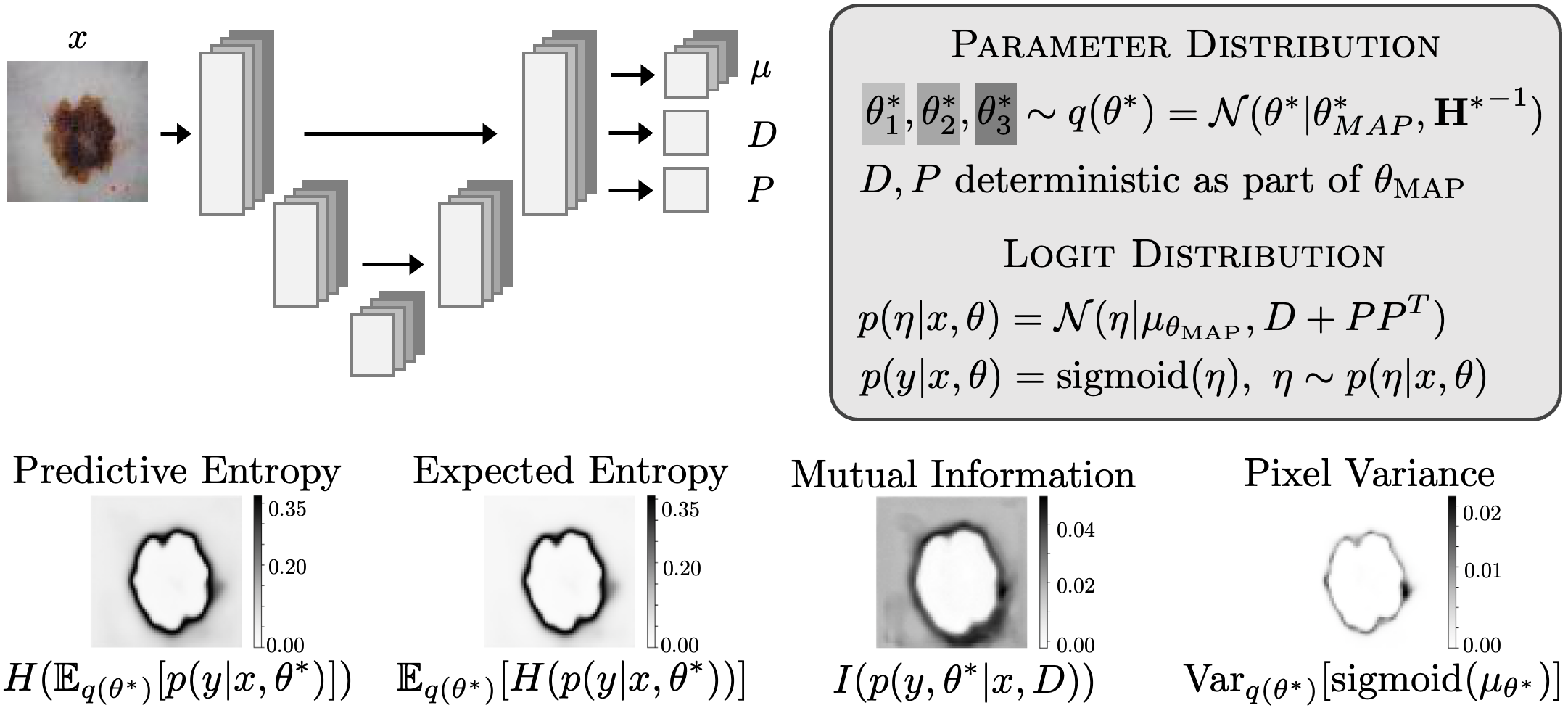}
%\vspace*{-\baselineskip}
\end{center}
   \caption{Model overview - uncertainty measures are calculated by approximating expectations by Monte Carlo-sampling mean networks from the Laplace approximation $q(\theta^*)$ and predicting the respective logit distributions $p(\eta \vert x, \theta)$ for $x$. }
\label{fig:model_overview}
%\vspace*{-\baselineskip}
\end{figure*}

The vectors $\theta_1 \in \Theta_1 = \mathbb{R}^T$ and $\theta_2 \in \Theta_2 = \mathbb{R}^T$ parameterize the mean and variance networks (c.f.\@ Eq.~\ref{logit_dist}) and share the first $t$ entries, i.e.we define the shared weight vector $\theta_t$ of the network by 
\begin{equation}
    \theta_{t} \coloneqq (\theta_{1_1}, \ldots ,\theta_{1_t} ) = (\theta_{2_1}, \ldots ,\theta_{2_t} ) \in \Theta_t =\mathbb{R}^t.
\end{equation}
Then $\theta \in \Theta = \mathbb{R}^{(t+2\cdot(T-t))}$ contains all model parameters
\begin{equation}
    \theta \coloneqq (\theta_t, \theta_{1_{t+1}}, \ldots, \theta_{1_{T}}, \theta_{2_{t+1}}, \ldots, \theta_{2_{T}}).
\end{equation}
The post-hoc Laplace approximation first finds a mode $\theta_{\MAP}$ by minimizing 
\begin{multline}
    \mathcal{L}(\theta) = - \log \mathbb{E}_{p(\eta \vert x, \theta)} [p(y \vert \eta)] - \log p(\theta) \approx  \\-\text{logsumexp}_{m=1}^M \left(\sum_{s=1}^S \log p(y_s \vert \eta_s^{(m)})\right) + \log(M) ,
\end{multline}
where $M$ logits $\eta$ are sampled from the distribution in Eq.~\eqref{logit_dist} and where the term $\log p(\theta)$ vanishes assuming a flat prior $\nabla_{\theta} p(\theta) = 0$. 
Since current algorithms for fast Hessian computations have no implementation for this loss function, we instead make use of the shared weights in the parameter vectors to estimate the mean and variance of the logit distribution based on the feature maps of a deep deterministic segmentation model. Using only one convolutional layer each for mean and variance estimation, we omit the entries of the variance heads on the parameter vector $\theta_{\MAP}$, i.e.\@ we set 
\begin{equation}
   \theta_{\MAP}^* \coloneqq \restr{\theta_{\MAP}}{(\theta_t, \theta_{1_{t+1}}, ..., \theta_{1_{T}})} \in \Theta_{\textrm{mean}} =\mathbb{R}^{T}.
\end{equation}
We can make use of the fact that the SSN loss function reduces to the binary cross entropy loss under zero variance, which allows us to fall back on the fast Hessian computation frameworks available. The posterior is then found by Laplace's method resulting in a Gaussian approximation in the parameter space $\Theta_{\textrm{mean}}$
\begin{equation}
    q(\theta^{*})= \mathcal{N}\left(\theta^{*} \vert \theta_{\MAP}^{*}, \mathbf{H^{*}}^{-1}\right), 
\end{equation}
with $\mathbf{H^*}$ defined as 
$
    \mathbf{H^*} = - \nabla_{\theta^{*}} \nabla_{\theta^{*}} \log \restr{p(\theta^{*} \vert D)} {\theta^{*} = \theta_{\MAP}^{*}}.
$
During inference we can now sample segmentation networks from the posterior distribution in form of the Laplace approximation. Each sampled segmentation network predicts one logit distribution.  Figure~\ref{fig:model_overview} gives an schematic overview of the proposed Laplacian Segmentation Network (LSN) and derived uncertainty measures. 
\subsection{Fast Hessian Approximations for Segmentation Networks with Skip Connections}\label{sec:fastH}
Computation of second order derivatives for Segmentation Networks is expensive due to the vast amount of parameters and pixels in the output. Standard methods approximate the Hessian with the diagonal of the Generalized Gauss Newton (\textsc{ggn}) matrix \cite{foresee1997ggn,botev2020}. This approximation, besides enforcing positive definiteness, also allows for an efficient backpropagation-like algorithm. The required compute scales linearly in the number of parameters and quadratic in the number pixels. The quadratic dependency is prohibitive already with images of size $64 \times 64$. We therefore make use of the diagonal backpropagation ($\textsc{db}$) proposed by \cite{miani_2022_neurips}, which returns a trace-preserving approximation of the diagonal of the \textsc{ggn}. The complexity of this approximation scales linearly with the number of pixels, allowing the computation of the Hessian also for larger images. The idea is to add a diagonal operator $\mathcal{D}$ in-between each backpropagation step. For each layer $l$
\begin{align}
    [\nabla_{\theta} & \nabla_{\theta} \log p(\theta \vert D)]_l
    \! \overset{\textsc{ggn}}{\approx} \!
    [J_\theta f_\theta (x)^\top \textbf{H}^{(L)} J_\theta f_\theta(x)]_l = \\ \nonumber
    & =
    J_\theta {f^{(l)}}^\top
        \left(
        \prod_{i=l+1}^L J_x {f^{(i)}}^\top
        \textbf{H}^{(L)}
        \prod_{i=L}^{l+1} J_x f^{(i)}
        \right)
    J_\theta f^{(l)} \\ \nonumber
    & \overset{\textsc{db}}{\approx}
    J_\theta {f^{(l)}}^\top
        \mathcal{D}
        \left(
        J_x {f^{(l+1)}}^\top
        \mathcal{D}
            \left(
            \dots
            \right)
        J_x f^{(l+1)}
        \right)
    J_\theta f^{(l)}
\end{align}
where $J_\theta$ denotes the Jacobian and $\textbf{H}^{(L)}$ the Hessian of the binary cross entropy loss with respect to the logits. The Hessian matrix can be expressed in closed form as a diagonal matrix plus an outer product matrix.

Moreover, we extend the \texttt{StochMan} library \cite{software:stochman} with support for skip-connection layers. For a given submodule $f_\theta$, a skip-connection layer $\textsc{sc}_f$ concatenates the function with the identity, such that $\textsc{sc}_f(x) = (f_\theta(x), x)$. The Jacobian is then defined as $J_x \textsc{sc}_f(x) := (J_x f_\theta(x), \mathbb{I}_x) $. We utilize the block structure of the Jacobian matrix and efficiently backpropagate its diagonal only. With a recursive call on the submodule $f$, the backpropagation supports nested skip-connections, i.e.\@ when some submodules of $f$ are skip-connections as well. This unlocks the use of various curvature-based methods for segmentation architectures with skip connection in future research. For a technical description of the used Hessian approximation we refer to the supplementary material.
\section{Experiments}
Our method validation is based on the ValUES framework by \cite{kahl2024values} for evaluating segmentation uncertainty. All benchmark models are constructed by combining an aleatoric and an epistemic component, using the same U-net backbone architecture for comparability. As aleatoric components we consider mean predictions of a U-net and mean-variance predictions of SSN, with diagonal and low-rank covariance matrices. The epistemic components are implemented as Ensembles, MC-Dropout and our post-hoc Laplace approximation. For the nine \textit{method combinations} we calculate the following uncertainty measures: Predictive Entropy, Expected Entropy, Mutual Information, Expected Pairwise KL (EPKL) \cite{schweighofer2023} and Pixel Variance. With exception of the EPKL all measures yield pixel-wise uncertainty heatmaps. We consider sum aggregation as well as a patch based strategy to take the uncertainty from pixel to image level. Patch aggregation sums uncertainties within a $10^2$ sliding window across the image, selecting the patch with the highest uncertainty as the image-level score. Table \ref{table_measures} lists the calculated uncertainty measures along their targeted uncertainty type and their definition.

\begin{figure*}[t]
    \begin{center}
    \includegraphics[width=0.9\linewidth]{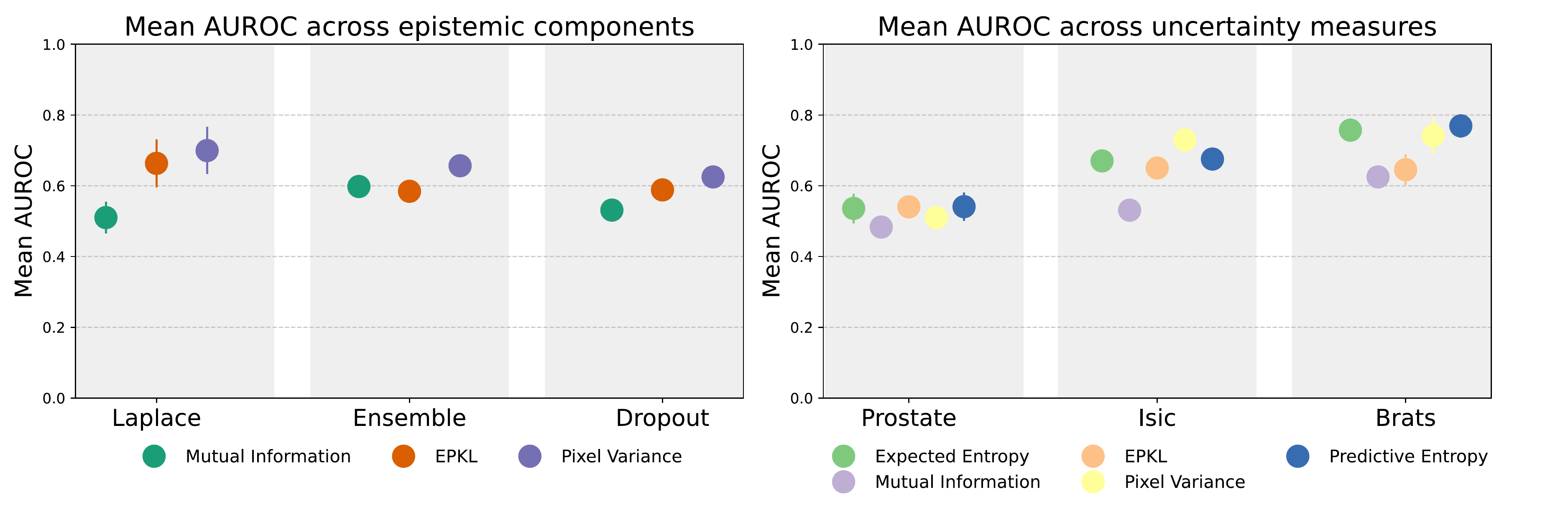}
    %\vspace*{-\baselineskip}
    %\vspace*{-\baselineskip}
    \end{center}
       \caption{\textbf{Left} - OOD performance measured by AUROC across epistemic components for Mutual Information (MI), Expected Pairwise KL (EPKL) and Pixel Variance (PV). Models using Laplace Approximations with EPKL and PV reach highest AUROC values on average. \textbf{Right} - Uncertainty Measures for predictive and aleatoric uncertainty perform on par indicating weak disentanglement.   }
    \label{fig:auroc_discussion}
    \vspace*{-\baselineskip}
\end{figure*}

All experiments are conducted on three datasets: the ISIC19 skin lesion dataset \cite{combalia2019bcn20000,codella2018skin,tschandl2018ham10000}, the BRATS dataset \cite{menze2014multimodal,bakas2017advancing,bakas2018identifying} and the first Prostate segmentation task from the QUBIQ 2021 challenge\footnote{https://qubiq21.grand-challenge.org}.
For the ISIC19 dataset we assign three OOD datasets, representing distribution shifts: Derm-Skin (DERM), Clin-Skin (CLINIC) \cite{Pacheco2020} and the PAD-UFES-20 dataset \cite{pacheco2020pad}. The Derm-Skin dataset contains 1,565 images of healthy skin, cropped out of the ISIC dataset. The Clin-Skin dataset contains 723 images showing healthy skin gathered from social networks. The PAD-UFES-20 dataset contains 1570 photos of skin lesions collected from smartphone cameras. For the BRATS and Prostate datasets, we follow the experimental setup of \cite{fuchs2021practical} and augment the images with Motion, Spike, Ghosting and Noise artifacts, which are regularly observed in MR images. For training and implementation details we refer to the supplementary material.

% \subsection{Out-of-distribution detection on image level}

\begin{figure*}[t]
    \begin{center}
    \includegraphics[width=0.8\linewidth]{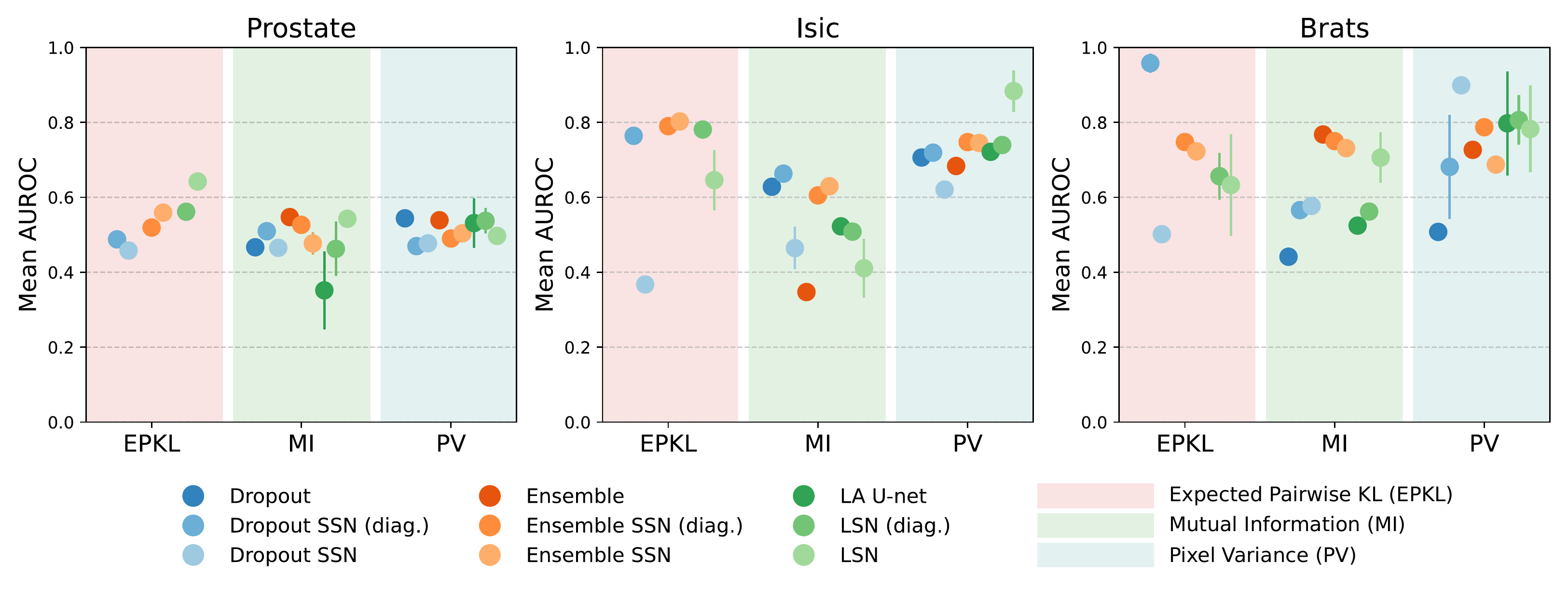}
    \end{center}
       \caption{\textbf{OOD performance} measured by AUROC across models, marginalized over aggregation strategies, for Mutual Information (MI), Expected Pairwise KL (EPKL) and Pixel Variance (PV). LSN models with EPKL and PV reach highest AUROC values for Prostate and ISIC respectively. }
    \label{fig:auroc_experiment}
    %\vspace*{-\baselineskip}
\end{figure*}

Image Level OOD detection can be viewed as a binary classification task across the ID test set and all OOD test sets. We therefore calculate the Area Under the Receiver Operating Characteristic Curve (AUROC) for all method combinations, uncertainty measures and aggregation strategies to assess how well different combinations can separate OOD from ID images. The AUROCs were calculated with \texttt{sklearn} \cite{scikit-learn} using per image ground truth binary labels (0-ID, 1-OOD) versus uncertainty scores as target predictions.
Figure~\ref{fig:auroc_experiment} shows the mean AUROC values for different method combinations and uncetainty measures across datasets. Note that for the method combinations, which use a U-net as an aleatoric component, the EPKL is not defined since it uses a KL divergence term, which is again not defined for two Dirac measures. For the Prostate and ISIC dataset, combinations of the LSN with EPKL and PV yield the highest AUROC values. On the BRATS dataset method combinations that use the Laplace approximation range after models using Dropout. We find that over all datasets method combinations that use the Laplace approximation for their parameter distribution yield the best OOD capabilities when combined with Pixel Variance and EPKL as shown in the left plot of Figure~\ref{fig:auroc_discussion}.

Additionally, we evaluate how well certain distribution shifts are detected by calculating the mean epistemic uncertainty assigned to a given OOD test set and normalizing it by the uncertainty assigned to the respective ID test set obtained from our ID dataset split. This compares how different distribution shifts influence the absolute values of epistemic uncertainty assigned, providing intuition on each method's sensitivity to OOD data. In Table \ref{uncertainty_ratios} we display which method combinations on the EPKL measure are able to assign 5\% and 10\% more uncertainty to the OOD set. Since the EPKL measure does not require aggregation over pixels, we can compare the method combinations directly on all datasets without bias introduced by a aggregation strategy. We find that the LSN model can detect most distribution shifts in terms of assigning a higher EPKL. The method, however, fails to identify Random Motion augmentations as ID, which have been used to augment images during training. 

%\vspace*{-\baselineskip}
\begin{table}
\scriptsize
\begin{center}
  \caption{OOD datasets with uncertainty measured by EPKL compared to ID datasets. A orange check (\greencheck) signifies an average uncertainty that is at least 5\% higher than the ID dataset, while a green check (\yellowcheck) indicates a 10\% or greater increase in uncertainty. For the ID Motion dataset an orange check (\greencheck \textcolor{BurntOrange}{*}) signifies an average uncertainty that is at most 5\% lower than the ID dataset. }
\label{uncertainty_ratios}
   %\vspace{.5em}
  \centering
  \begin{tabular}{lllll|llll|lll|ll}
    \toprule
    
    & \multicolumn{4}{c}{Prostate} & \multicolumn{4}{c}{Brats}& \multicolumn{3}{c}{ISIC}\\
    \cmidrule{2-5}
    \cmidrule{6-9}
    \cmidrule{10-12}

    Model       & Motion & Ghost & Spike & Blur & Motion & Ghost & Spike & Blur & Clin & Derm & Padufes &\greencheck &\yellowcheck \\
    \midrule
    Ensemble SSN (diag.)  & \xmark  & \greencheck \yellowcheck & \greencheck \yellowcheck & \xmark & \xmark & \greencheck \yellowcheck & \greencheck \yellowcheck& \xmark & \greencheck \yellowcheck & \greencheck \yellowcheck  & \greencheck \yellowcheck & 7& 7\\
    Ensemble SSN  & \xmark  & \greencheck \yellowcheck & \greencheck & \xmark & \xmark & \greencheck \yellowcheck & \greencheck \yellowcheck  & \greencheck \yellowcheck & \greencheck \yellowcheck & \greencheck \yellowcheck  & \greencheck \yellowcheck &8&7\\
    \midrule
    Dropout SSN (diag.)       & \xmark  & \greencheck \yellowcheck& \greencheck \yellowcheck& \xmark & \xmark & \greencheck \yellowcheck  & \greencheck \yellowcheck & \xmark & \greencheck \yellowcheck & \greencheck \yellowcheck  &\greencheck \yellowcheck &7&7\\
    Dropout SSN      & \greencheck \textcolor{BurntOrange}{*}  & \greencheck \yellowcheck& \greencheck \yellowcheck& \xmark & \xmark & \greencheck \yellowcheck  & \greencheck \yellowcheck & \xmark & \xmark & \xmark  & \greencheck \yellowcheck &5&6\\
    \midrule 
    LSN (diag.)    & \greencheck \textcolor{BurntOrange}{*} & \greencheck & \xmark & \xmark & \xmark & \greencheck \yellowcheck  & \greencheck \yellowcheck & \xmark & \greencheck \yellowcheck & \greencheck \yellowcheck  & \greencheck \yellowcheck &7&5\\
    \textbf{LSN}      & \xmark  & \greencheck \yellowcheck & \greencheck \yellowcheck & \greencheck & \xmark & \greencheck \yellowcheck & \greencheck \yellowcheck & \greencheck \yellowcheck & \greencheck \yellowcheck & \greencheck \yellowcheck  & \greencheck \yellowcheck &\textbf{9}&\textbf{8}\\
    \bottomrule
  \end{tabular}

   %\vspace{.5em}
  \centering
 
%\vspace*{-\baselineskip}
%\vspace*{-\baselineskip}
\end{center}
\end{table}
%\vspace*{-\baselineskip}
%\vspace*{-\baselineskip}

\section{Discussion and Conclusion}
In this paper, we have demonstrated how Laplace approximations can scale to image segmentation tasks, through a trace-preserving diagonal Hessian approximation. Importantly, this scales linearly with the number of image pixels, unlike past work which exhibited a quadratic complexity. We have demonstrated across different datasets that the parameter distributions obtained by Laplace's method, in combination with suitable uncertainty measures, can lead to superior OOD detection performance on image level. 

Our experimental findings support the recent initiative in research for finding better measures for epistemic uncertainty than Mutual Information \cite{schweighofer2023}. Marginalizing over all datasets and aggregations strategies, our findings show that EPKL and Pixel Variance, not Mutual Information, provide the strongest discriminative power for classifying images as either ID or OOD (cf. Fig. \ref{fig:auroc_discussion}, left). Further we find that there is still a strong correlation between aleatoric and epistemic measures across all method combinations, visible by comparable AUROC performance over all datasets(cf. Fig. \ref{fig:auroc_discussion}, right).  

Further research might investigate in more depth how different logit distributions interplay with Laplace approximations in theoretically suggested uncertainty measures that target aleatoric and epistemic uncertainty. The presented method provides an extendable framework for other researchers to build upon.

\begin{credits}
\subsubsection{\ackname}  
This work was supported by VILLUM FONDEN (grants 15334, 42062), the European Research Council under the European Union's Horizon 2020 research and innovation programme (grant 757360), Novo Nordisk Foundation (NNF20O-C0062606), LANL (LA-UR-24-23937) LDRD grant 20210043DR (U.S. DOE NNSA Contract 89233218CNA000001), and the Pioneer Centre for AI (DNRF grant P1). 

\subsubsection{\discintname}
The authors have no competing interests to declare that are
relevant to the content of this article.

\end{credits}

{
\bibliographystyle{splncs04}
\bibliography{references}
}

\appendix

\title{Supplementary to Laplacian Segmentation Networks Improve Epistemic Uncertainty Quantification}
\titlerunning{Supplementary Laplacian Segmentation Networks}
\author{}
\authorrunning{F. Author et al.}
\institute{}
\maketitle              % typeset the header of the contribution
\vspace*{-\baselineskip}
\vspace*{-\baselineskip}

\section{Implementation and Training Details}

Training configurations are provided in Table \ref{training_configs}. All models were trained with the Adam optimizer.  The U-net backbone was constructed with feature maps of size $8, 16, 32, 64, 128$. Uncertainty measures were approximated from $50$ samples from the posterior and $20$ samples from the logit distribution. 

\vspace*{-\baselineskip}
\begin{table}
% \scriptsize
\begin{center}
  \caption{Implementation and Training Details.  Dropout models for the ISIC dataset were trained with a $0.0005$ learning rate to improve convergence. }
\label{training_configs}
   \vspace{.5em}
  \centering
  \begin{tabular}{p{30mm}p{20mm}|p{20mm}|p{20mm}}
    \toprule
    
    & \multicolumn{3}{c}{Dataset}\\
    \cmidrule{2-4}
    % \cmidrule{6-9}
    % \cmidrule{10-12}

    Configuration       & ISIC & Prostate & Brats \\
    \hline
   Epochs & 60 & 150 & 600 \\
   \hline
    Batch Size  & 32 & 10 & 32  \\
    \hline
    Learning Rate   & 0.001* & 0.001    & 0.001  \\
    \bottomrule
  \end{tabular}

   \vspace{.5em}
  \centering
 
\vspace*{-\baselineskip}
\vspace*{-\baselineskip}
\end{center}
\end{table}
\vspace*{-\baselineskip}

\section{Fast Hessian Approximation}

Consider a neural network (\textsc{nn}) $f_\theta:\mathcal{X}\rightarrow\mathcal{Y}$ with $L$ layers. The parameter $\theta = (\theta_1, \dots, \theta_L) \in\Theta$ is the concatenation of the parameters for each layer $i \in \{1,...,L \}$. The \textsc{nn} $f_\theta=
f^{(L)}_{\theta_L}\circ f^{(L-1)}_{\theta_{L-1}} 
\circ\,\dots\,\circ 
f^{(2)}_{\theta_2} \circ f^{(1)}_{\theta_1}$ is a composition of $L$ functions $f^{(L)},f^{(L-1)},\dots,f^{(1)}$, where $f^{(i)}$ is parametrized by $\theta_{i}$. Let $x_0\in\mathcal{X}$ be the input and  $x_i:=f^{(i)}_{\theta_i}(x_{i-1})$ for $i=1,\dots,L$, such that the \textsc{nn} output is $x_L\in\mathcal{Y}$. We define the \emph{diagonal} operator $\mathcal{D}: 
\mathbb{R}^{m\times m}
\rightarrow
\mathbb{R}^{m\times m} $ on quadratic matrices as
\[
[\mathcal{D}(M)]_{ij}
:=
\left\{
\begin{array}{ll}
    M_{ij} & \text{ if } i=j \\
    0 & \text{ if } i\not=j
\end{array}
\right.
\qquad
\forall i,j=1,\dots,m.
\]
The \textbf{Jacobian} $J_\theta f_\theta(x_0)$ of the \textsc{nn} %w.r.t.\@ the parameter 
has a layer block structure, block $i$ is %$\theta_i$ of layer $i$ as
\begin{equation*}\label{eq:jacobian_chain_rule}
J_{\theta_i}f_{\theta}(x_0)
=
J_{\theta_i}
\left(
        f^{(i)}_{\theta_i}
        \circ\dots\circ
        f^{(L)}_{\theta_L}
    \right)
(x_{i-1}) 
=
\left(
    \prod_{j=L}^{i+1} 
    J_{x_{j-1}}f^{(j)}_{\theta_j}(x_{j-1})
\right)
J_{\theta_i}f^{(i)}_{\theta_i}(x_{i-1}).
\end{equation*}

The Laplace approximation requires the Hessian \textbf{H} of the loss w.r.t. the parameters $\nabla^2_{\theta} \mathcal{L}(f_\theta(x_0))
\in\mathbb{R}^{|\theta|\times|\theta|}$. Using the chain rule it holds, that
\begin{equation*}\label{eq:hessian_chain_rule}
\underbrace{\nabla^2_{\theta} \mathcal{L}(f_\theta(x_0))}_{=:H_\theta}
=
\underbrace{
J_{\theta}f_\theta(x_0)^\top
    \cdot 
    \nabla^2_{x_L}\mathcal{L}(x_L) 
    \cdot
J_{\theta}f_\theta(x_0)
}_{=:\textsc{ggn}_\theta}
+
\sum_{o=1}^{|x_L|} 
    [\nabla_{x_L}\mathcal{L}(x_L)]_o \cdot
    \nabla^2_{\theta} [f_\theta(x_0)]_o,
\end{equation*}
where $[v]_o$ refers to the $o$-th component of vector $v$ and $|v|$ to its length. We can write the diagonal block $\textsc{ggnb}^{(i)}_\theta=J_{\theta_i}f_\theta(x_0)^\top \textbf{H}^{\mathcal{L}} J_{\theta_i}f_\theta(x_0)$ of the $i$-th layer as

\begin{equation}\label{eq:ggn_chain_rule}
\begin{aligned}
    \textsc{ggnb}^{(i)}_\theta
    = &
    J_{\theta_i}f_{\theta}(x_0)^\top \cdot \textbf{H}^{\mathcal{L}} \cdot J_{\theta_i}f_{\theta}(x_0)
\end{aligned}
\end{equation}

From this expression, plus the chain rule expansion of the Jacobian, we can build an efficient backpropagation-like algorithm to compute $\textsc{ggnb}_\theta$, it start from $\textbf{H}^{\mathcal{L}}$ and then iterated backward over layers. The same holds for the diagonal approximation, which we refer to as $\textsc{ggnd}_\theta:=\mathcal{D}(\textsc{ggn}_\theta)=\mathcal{D}(\textsc{ggnb}_\theta)$. This approach already scales linearly in the number of parameter $|\theta|$.
On top of that, the diagonal backpropagation approximates the diagonal of the Generalized Gauss-Newton matrix. It is defined, for each layer $i$, by adding a diagonalization operator in between each Jacobian product, marked \textcolor{BrickRed}{red} in Algorithm 1. Without this extra operator the algorithm would return the exact diagonal.

\vspace*{-\baselineskip}
\begin{algorithm}[H]
\caption{Computation of $\textsc{db}_\theta$}\label{alg:exact_backprop_diag}
\begin{algorithmic}
\State $M$ = $\textbf{H}^{\mathcal{L}}$
\For{$j=L,L-1,\dots,1$}
\State $\textsc{db}^{(j)}_\theta$ = $\mathcal{D}\left(J_{\theta_j}f^{(j)}_{\theta_j}(x_{j-1})^\top \cdot M \cdot J_{\theta_j}f^{(j)}_{\theta_j}(x_{j-1})\right)$
\State $M$ = $\textcolor{BrickRed}{\mathcal{D}}\textcolor{BrickRed}{\Big(} J_{x_{j-1}}f^{(j)}_{\theta_j}(x_{j-1})^\top \cdot M \cdot J_{x_{j-1}}f^{(j)}_{\theta_j}(x_{j-1})\textcolor{BrickRed}{\Big)}$
\EndFor
\State $\textsc{db}_\theta$ = $(\textsc{db}^{(1)}_\theta, \dots, \textsc{db}^{(L)}_\theta)$
\State \textbf{return} $\textsc{db}_\theta$
\end{algorithmic}
\end{algorithm}
\vspace*{-\baselineskip}

\textbf{Proposition.} For an autoencoder network, the memory requirement of the Algorithm scale \emph{linearly} both in number of parameter and in number of pixels.
\begin{proof}
    The bottlenecks are the storage of the matrixes $\textsc{db}^{(j)}_\theta
\in\mathbb{R}^{|\theta_j|}$ and $
M
\in\mathbb{R}^{|x_{j-1}|}
$ at each step $j$
\end{proof}

\paragraph{Skip-connections}
For any given submodule $g_\theta$, a skip-connection layer $\textsc{sc}(g)_\theta$ is defined as $x  \longmapsto (g_\theta(x), x)$. The Jacobian with respect to the parameter is the same as the Jacobian of the $g_\theta$ while the Jacobian with respect to the input is $
    J_x \textsc{sc}(g)_\theta(x) 
    =
    \left(\begin{array}{c}
        J_x g_\theta(x) \\ \hline
        \mathbb{I}
    \end{array}\right)
    \in\mathbb{R}^{(O+I)\times I}$. 

\textbf{Proposition.} If $M$ is diagonal, then one step of Alg 1 can be computed as
\begin{equation*}\label{eq:skipconn_recursive_def}
    \mathcal{D}\left( J_x \textsc{SC}(g)_\theta(x)^\top \cdot M \cdot J_x \textsc{SC}(g)_\theta(x) \right)
    =
    \mathcal{D}\left( J_x g_\theta(x)^\top M_{11} J_x g_\theta(x) \right)
    + \mathcal{D} (M_{22}).
\end{equation*}
\begin{proof}
    Let $
    M
    =
    \begin{pmatrix}
    M_{11} & M_{12} \\
    M_{21} & M_{22}
    \end{pmatrix}$ and then
    
\begin{align*}
    J_x \textsc{SC}(g)_\theta(x)^\top \cdot &\,M \cdot J_x \textsc{SC}(g)_\theta(x) 
     =
    \left(\begin{array}{c|c}
        J_x g_\theta(x)^\top &
        \mathbb{I}
    \end{array}\right)
    \begin{pmatrix}
    M_{11} & M_{12} \\
    M_{21} & M_{22}
    \end{pmatrix}
    \left(\begin{array}{c}
        J_x g_\theta(x) \\ \hline
        \mathbb{I}
    \end{array}\right) \\
    & =
    J_x g_\theta(x)^\top M_{11} J_x g_\theta(x)
    + M_{12} J_x g_\theta(x)
    + J_x g_\theta(x)^\top M_{21}
    + M_{22}
\end{align*}
\end{proof}


\begin{comment}

{\small
\bibliographystyle{splncs04}
\bibliography{references}

\begin{thebibliography}{10}
\providecommand{\url}[1]{\texttt{#1}}
\providecommand{\urlprefix}{URL }
\providecommand{\doi}[1]{https://doi.org/#1}

\bibitem{bakas2017advancing}
Bakas, S., et~al.: Advancing the cancer genome atlas glioma mri collections
  with expert segmentation labels and radiomic features. Scientific data
  \textbf{4}(1),  1--13 (2017)

\bibitem{bakas2018identifying}
Bakas, S., et~al.: Identifying the best machine learning algorithms for brain
  tumor segmentation, progression assessment, and overall survival prediction
  in the brats challenge. arXiv preprint arXiv:1811.02629  (2018)

\bibitem{baumgartner2019phiseg}
Baumgartner, C.F., et~al.: {PHiSeg}: Capturing uncertainty in medical image
  segmentation. Medical Image Computing and Computer Assisted Intervention
  (MICCAI) pp. 119--127 (2019)

\bibitem{bishop2006}
Bishop, C.M.: Pattern Recognition and Machine Learning (Information Science and
  Statistics). Springer-Verlag, Berlin, Heidelberg (2006)

\bibitem{botev2020}
Botev, A.: The Gauss-Newton matrix for deep learning models and its
  applications. Ph.D. thesis, UCL (University College London) (2020)

\bibitem{codella2018skin}
Codella, N.C., et~al.: Skin lesion analysis toward melanoma detection: A
  challenge at the 2017 international symposium on biomedical imaging {(ISBI)},
  hosted by the international skin imaging collaboration ({ISIC}). In: IEEE
  15th International Symposium on Biomedical Imaging. pp. 168--172. IEEE (2018)

\bibitem{combalia2019bcn20000}
Combalia, M., et~al.: {BCN20000}: Dermoscopic lesions in the wild. arXiv
  preprint arXiv:1908.02288  (2019)

\bibitem{daxberger2021laplaceredux}
Daxberger, E., Kristiadi, A., Immer, A., Eschenhagen, R., Bauer, M., Hennig,
  P.: Laplace redux--effortless {B}ayesian deep learning. In: {N}eur{IPS}
  (2021)

\bibitem{software:stochman}
Detlefsen, N.S., Pouplin, A., Feldager, C.W., Geng, C., Kalatzis, D.,
  Hauschultz, H., González-Duque, M., Warburg, F., Miani, M., Hauberg, S.:
  Stochman. GitHub. Note:
  https://github.com/MachineLearningLifeScience/stochman/  (2021)

\bibitem{foresee1997ggn}
Foresee, F.D., Hagan, M.T.: Gauss-{N}ewton approximation to {B}ayesian
  learning. In: Proceedings of International Conference on Neural Networks
  (ICNN'97). vol.~3, pp. 1930--1935. IEEE (1997)

\bibitem{fuchs2021practical}
Fuchs, M., Gonzalez, C., Mukhopadhyay, A.: Practical uncertainty quantification
  for brain tumor segmentation. In: Medical Imaging with Deep Learning (2021)

\bibitem{gawlikowski2021}
Gawlikowski, J., et~al.: A survey of uncertainty in deep neural networks. arXiv
  preprint arXiv:2107.03342  (2021)

\bibitem{hendrycks2016baseline}
Hendrycks, D., Gimpel, K.: A baseline for detecting misclassified and
  out-of-distribution examples in neural networks. arXiv preprint
  arXiv:1610.02136  (2016)

\bibitem{houlsby2011bayesian}
Houlsby, N., Huszár, F., Ghahramani, Z., Lengyel, M.: Bayesian active learning
  for classification and preference learning (2011)

\bibitem{kahl2024values}
Kahl, K.C., L{\"u}th, C.T., Zenk, M., Maier-Hein, K., Jaeger, P.F.: Values: A
  framework for systematic validation of uncertainty estimation in semantic
  segmentation. arXiv preprint arXiv:2401.08501  (2024)

\bibitem{kendall2017}
Kendall, A., Gal, Y.: What uncertainties do we need in {Bayesian} deep learning
  for computer vision? Advances in Neural Information Processing Systems
  \textbf{30} (2017)

\bibitem{ditlevsen2008}
Kiureghian, A.D., Ditlevsen, O.: Aleatory or epistemic? {Does} it matter?
  Structural Safety  \textbf{31}(2),  105--112 (2009)

\bibitem{kohl2018probunet}
Kohl, S., et~al.: A probabilistic {U-Net} for segmentation of ambiguous images.
  Advances in Neural Information Processing Systems  \textbf{31} (2018)

\bibitem{kohl2019hierarchical}
Kohl, S.A.A., et~al.: A hierarchical probabilistic {U-Net} for modeling
  multi-scale ambiguities. arXiv preprint arXiv:1905.13077  (2019)

\bibitem{blundell2016ensemble}
Lakshminarayanan, B., Pritzel, A., Blundell, C.: Simple and scalable predictive
  uncertainty estimation using deep ensembles. Advances in Neural Information
  Processing Systems  \textbf{30} (2017)

\bibitem{lee2016stochastic}
Lee, S., Purushwalkam Shiva~Prakash, S., Cogswell, M., Ranjan, V., Crandall,
  D., Batra, D.: Stochastic multiple choice learning for training diverse deep
  ensembles. Advances in Neural Information Processing Systems  \textbf{29}
  (2016)

\bibitem{mackay1992bayesian}
MacKay, D.J.: Bayesian interpolation. Neural computation  \textbf{4}(3),
  415--447 (1992)

\bibitem{menze2014multimodal}
Menze, B.H., et~al.: The multimodal brain tumor image segmentation benchmark
  (brats). IEEE transactions on medical imaging  \textbf{34}(10),  1993--2024
  (2014)

\bibitem{miani_2022_neurips}
Miani, M., Warburg, F., Moreno-Mu{\~n}oz, P., Detlefsen, N.S., Hauberg, S.:
  Laplacian autoencoders for learning stochastic representations. In: Advances
  in Neural Information Processing Systems (2022)

\bibitem{monteiro2020}
Monteiro, M., et~al.: Stochastic segmentation networks: Modelling spatially
  correlated aleatoric uncertainty. Advances in Neural Information Processing
  Systems  \textbf{33},  12756--12767 (2020)

\bibitem{mucsanyi2024benchmarking}
Mucsányi, B., Kirchhof, M., Oh, S.J.: Benchmarking uncertainty
  disentanglement: Specialized uncertainties for specialized tasks (2024)

\bibitem{Pacheco2020}
Pacheco, A.G.C., Sastry, C.S., Trappenberg, T., Oore, S., Krohling, R.A.: On
  out-of-distribution detection algorithms with deep neural skin cancer
  classifiers. In: 2020 IEEE/CVF Conference on Computer Vision and Pattern
  Recognition Workshops (CVPRW). pp. 3152--3161 (2020).
  \doi{10.1109/CVPRW50498.2020.00374}

\bibitem{pacheco2020pad}
Pacheco, A.G.h.o.: Pad-ufes-20: A skin lesion dataset composed of patient data
  and clinical images collected from smartphones. Data in brief  \textbf{32},
  106221 (2020)

\bibitem{scikit-learn}
Pedregosa, F., Varoquaux, G., Gramfort, A., Michel, V., Thirion, B., Grisel,
  O., Blondel, M., Prettenhofer, P., Weiss, R., Dubourg, V., Vanderplas, J.,
  Passos, A., Cournapeau, D., Brucher, M., Perrot, M., Duchesnay, E.:
  Scikit-learn: Machine learning in {P}ython. Journal of Machine Learning
  Research  \textbf{12},  2825--2830 (2011)

\bibitem{popescu2021distributional}
Popescu, S.G., Sharp, D.J., Cole, J.H., Kamnitsas, K., Glocker, B.:
  Distributional {G}aussian process layers for outlier detection in image
  segmentation. In: International Conference on Information Processing in
  Medical Imaging. pp. 415--427. Springer (2021)

\bibitem{ronneberger2015}
Ronneberger, O., Fischer, P., Brox, T.: U-net: Convolutional networks for
  biomedical image segmentation. CoRR  \textbf{abs/1505.04597} (2015),
  \url{http://arxiv.org/abs/1505.04597}

\bibitem{rupprecht2017learning}
Rupprecht, C., Laina, I., DiPietro, R., Baust, M., Tombari, F., Navab, N.,
  Hager, G.D.: Learning in an uncertain world: Representing ambiguity through
  multiple hypotheses. In: Proceedings of the IEEE International Conference on
  Computer Vision. pp. 3591--3600 (2017)

\bibitem{schweighofer2023}
Schweighofer, K., Aichberger, L., Ielanskyi, M., Hochreiter, S.: Introducing an
  improved information-theoretic measure of predictive uncertainty  (2023),
  \url{https://openreview.net/forum?id=c71B6zW70d}

\bibitem{schweighofer2023quantification}
Schweighofer, K., Aichberger, L., Ielanskyi, M., Klambauer, G., Hochreiter, S.:
  Quantification of uncertainty with adversarial models. In: Thirty-seventh
  Conference on Neural Information Processing Systems (2023),
  \url{https://openreview.net/forum?id=5eu00pcLWa}

\bibitem{selvan2020uncertainty}
Selvan, R., Faye, F., Middleton, J., Pai, A.: Uncertainty quantification in
  medical image segmentation with normalizing flows. In: Machine Learning in
  Medical Imaging. Lecture Notes in Computer Science, Springer, Switzerland
  (2020)

\bibitem{tschandl2018ham10000}
Tschandl, P., Rosendahl, C., Kittler, H.: The {HAM10000} dataset, a large
  collection of multi-source dermatoscopic images of common pigmented skin
  lesions. Scientific Data  \textbf{5}(1), ~1--9 (2018)

\bibitem{pmlr-v216-wimmer23a}
Wimmer, L., Sale, Y., Hofman, P., Bischl, B., H\"ullermeier, E.: Quantifying
  aleatoric and epistemic uncertainty in machine learning: Are conditional
  entropy and mutual information appropriate measures? In: Evans, R.J.,
  Shpitser, I. (eds.) Proceedings of the Thirty-Ninth Conference on Uncertainty
  in Artificial Intelligence. Proceedings of Machine Learning Research,
  vol.~216, pp. 2282--2292. PMLR (31 Jul--04 Aug 2023),
  \url{https://proceedings.mlr.press/v216/wimmer23a.html}

\end{thebibliography}
}

\end{comment}

\end{document}